# KAN based Autoencoders for Factor Models

Tianqi Wang
Courant Institute of Mathematical Science
New York University
New York, NY, US
tw2250@nyu.edu

Shubham Singh
ECE Department, NYU Tandon School of Engineering
New York University
New York, NY, US

## ABSTRACT

Inspired by recent advances in Kolmogorov-Arnold Networks (KANs), we introduce a novel approach to latent factor conditional asset pricing models. While previous machine learning applications in asset pricing have predominantly used Multilayer Perceptrons with ReLU activation functions to model latent factor exposures, our method introduces a KAN-based autoencoder which surpasses MLP models in both accuracy and interpretability. Our model offers enhanced flexibility in approximating exposures as nonlinear functions of asset characteristics, while simultaneously providing users with an intuitive framework for interpreting latent factors. Empirical backtesting demonstrates our model's superior ability to explain cross-sectional risk exposures. Moreover, long-short portfolios constructed using our model's predictions achieve higher Sharpe ratios, highlighting its practical value in investment management.

## KEYWORDS

Deep learning, Autoencoder, Kolmogorov-Arnold Networks (KANs), Asset pricing models, Non-linear factor models, Predictive modeling, Quantitative finance

## CCS CONCEPTS

• Computing methodologies: Machine learning algorithms. • Applied computing: Finance. • Theory of computation: Neural networks.

## 1 INTRODUCTION

Factor models are fundamental in quantitative finance. It provides a framework to forecast asset returns beyond standard return variation as a compensation for aggregated risk factors. In an economically intuitive way, the practitioners and portfolio managers conduct performance attribution and mean-variance-optimization based on asset's exposures to common risk factors. The most general form of factor model takes the form:

$$r_t = \alpha_t + \beta_t * factor_t + \epsilon_t$$

where, $r_t$ is $(n \times 1)$ cross sectional asset excess returns vector at time $t$, $\alpha_t$ is the $(n \times 1)$ intercept term, often interpreted as asset's abnormal returns. $\beta_t$ is the factor exposure matrix, where each row contains the asset's exposures to risk factors. $factor_t$ is the $(k \times n)$ risk factor returns. $\epsilon_t$ is the $(n \times 1)$ idiosyncratic return vector, representing asset-specific risk that is not captured by the market wide risk factors.

Past trends in academic literature have been focusing on identifying economically important risk factors [1][2][3]. But Traditional models like the CAPM and the Fama-French three-factor model rely on linear relationships between cross sectional asset returns with the returns of a set of predetermined characteristic portfolios. However, there is no intuitive reason for this simplifying assumption: There is no guarantee that our set of risk factors best explains the variation in cross sectional returns. Moreover, the relationship between these risk factors and asset returns may not necessarily be linear.

This issue is partially resolved with the Statistical Factor Model, which relies on PCA to find a set of latent factors that maximize the explained variance of returns. [5]. But the PCA-based model is well-known for its lack of interpretability and inefficient information use, as it is solely based on returns series and ignores additional asset specific characteristics.

In light of this limitation, the asset pricing models proposed by KPS incorporate an additional flexibility, which assumes a factor structure of the form:

$$r_t = \beta(Z_{t-1}) f_t + \epsilon_t$$

where factors $f_t$ are now treated as latent, and factor exposures are modeled as a function of $Z_{t-1}$, a $N * P$ characteristic matrix, where $P$ stands for the number of asset specific characteristics.

The functional form of β is a modeling choice. KPS assumes that the mapping from P characteristics to K latent factors is linear, i.e.

$$\beta(Z_{t-1}) = Z_{t-1} \Gamma$$

For some matrix Γ. Although this formulation has a particularly tractable estimation strategy for both latent factors and factor exposures, the level of flexibility brought by the linear model is not always satisfying.

Recent advances in deep learning have paved the way for non-linear models such as autoencoders, which have shown promise in capturing intricate relationships in financial data by removing noise, or modeling functions such that features can be linearly represented in factor models. [6] assumes a more general form of beta functions, and proposes to use neural networks to approximate it.

This paper builds on this foundation by introducing [7] Kolmogorov-Arnold Networks (KANs) for asset pricing models. KANs are particularly adept at representing nonlinear functions due to their ability to decompose complex functions into simpler ones and then combine these components to approximate the original function through use of learnable activation functions .

Traditional activation functions such as ReLU, GeLU, Sigmoid, Tanh, Maxout, although works well for some problems, can fail to approximate certain functions. For example, because ReLU activation functions introduce piecewise linearity to the model, it leads to inefficiency in approximating smooth functions like sine and cosine. Although switching the type of activation functions may sidestep this problem, for most applications, the optimal choice of activation functions is not immediately clear and can be difficult to decide.

In this study, we propose an implementation of a latent factor conditional asset pricing model using KANs within an autoencoder architecture. Unlike traditional methods that rely on Multilayer Perceptrons (MLPs) with ReLU activations, our approach harnesses the power of KANs to enhance both the accuracy and interpretability of latent factor exposures. By embedding KANs into the beta network of an autoencoder, we aim to provide a more robust noiseless framework for modeling complex relationships in financial data.

Integrating KANs into the beta network of an autoencoder for latent factor modeling and demonstrating superior predictive performance and interpretability through extensive empirical analysis.

The remainder of this paper is organized as follows: Section 2 reviews relevant literature on factor models and deep learning in finance. Section 3 details our methodology, focusing on the integration of KANs into the autoencoder framework. Section 4 presents empirical results, followed by a discussion in Section 5. Finally, Section 6 concludes with implications for future research and applications.

## 2 LITERATURE REVIEW

### 2.1 Kolmogorov-Arnold Networks

Kolmogorov-Arnold Networks represent a paradigm shift in neural network architecture, inspired by the Kolmogorov-Arnold representation theorem [8]. This theorem posits that any multivariate continuous function can be decomposed into a finite composition of continuous functions of a single variable, combined with addition operations [9]. KANs diverge from traditional neural networks by embedding learnable activation functions directly on the edges connecting nodes. Parameterized spline functions replace the typical weight matrices found in Multilayer Perceptrons (MLPs), enabling flexible and adaptive nonlinear transformations of input data. This design enhances the model's expressiveness and interpretability while maintaining efficiency [8].

The efficiency of KANs lies in their ability to represent complex functions using fewer parameters compared to traditional MLPs, crucial for applications involving high-dimensional data such as finance [10]. KANs mitigate the curse of dimensionality often encountered with spline-based models by leveraging the compositional nature of the Kolmogorov-Arnold theorem. The KAN architecture introduces the concept of a "KAN layer," allowing for deeper representations beyond the original two-layer formulation [8]. This ability to handle high-dimensional data efficiently while providing interpretable results makes KANs particularly promising for financial applications where transparency is crucial.

## 2.2 Autoencoders for Factor Models

Autoencoder models have advanced financial econometrics by improving factor models through noise removal. Recent innovations include the integration of variational autoencoders (VAEs) with dynamic factor models [11], the development of conditional autoencoder (CA) asset pricing models for specific markets [12], and the application of asymmetric autoencoders for covariance matrix estimation [13]. Deep learning methodologies have led to models integrating Beta and Factor networks, surpassing traditional linear frameworks in capturing the complexities of asset returns [14].

The application of KANs in autoencoder models is an attempt to create factor models that are both highly expressive and interpretable, addressing a key challenge in applying AI to finance. Future research may explore the applicability of these models to a wider range of financial instruments and market conditions, their performance in real-time decision-making scenarios, and methods to ensure their robustness in volatile markets. Addressing interpretability, regulatory compliance, and ethical considerations will be crucial for their widespread adoption in the financial industry.

## 3 METHODOLOGY

### 3.1 Dataset Description

We utilize the same data as [6] (GU 2020). We analyze the same dataset studied in Gu et al. (2019), which contains monthly individual stock returns from the Center for Research in Securities Prices (CRSP) for all firms listed in the three major exchanges: NYSE, AMEX, and NASDAQ. We use the Treasury bill rate to proxy for the risk-free rate from which we calculate individual excess returns. Our sample begins in March 1957 (the start date of the S&P 500) and ends in December 2016, totaling 60 years.

The dataset is prepared with a lag. We assume that the monthly stock characteristics are delayed for a month, quarterly stock characteristics are delayed for 4 months, and annual characteristics are delayed for a year. This practice avoids look-ahead bias in our simulation.

No filters are imposed on the stocks that are included in our simulation. The total number of stocks in our sample is nearly 30,000, with the average number of stocks per month exceeding 6,200.

### 3.2 Model Design

We deploy a similar model architecture as Gu et al. At the highest level, we assumes factor structure of the form:

$$r_t = \beta(Z_{t-1}) f_t + \epsilon_t$$

where the factor exposure $\beta$ is a general function of asset specific covariates, and $f_t$ is factor return at time t.

Our main difference from Gu et al is the choice of function approximator for $\beta$. Instead of a MLP, we use a KAN network to model the factor exposure. The core operational principle of KANs in our model involves embedding spline functions within the network's architecture. For each edge connecting nodes in the Beta network, a spline function is parameterized to adaptively transform the input data. This operation is mathematically represented as:

$$y_i = \sum_{j=1}^{n} w_{ij} * Spline(x_j)$$

Written down in matrix notation, this amounts to the following:

$$\beta(z_{i,t-1}) := KAN(z_{i,t-1}) = (\Phi_{L-1} \circ ... \circ \Phi_1 \circ \Phi_0) z_{i, t-1}$$

Where $\Phi_i$ is the ith layer of KAN network, defined as a shape $n_{in} \times n_{out}$ matrix of 1d spline function $\phi_{i,j}$ each with trainable parameters. In practice, it is empirically discovered that the KAN network works best in latent space, so we add embedding layers for both inputs and outputs in the form:

$$\beta(z_{i,t-1}) = \Gamma_{out} (\Phi_{L-1} \circ ... \circ \Phi_1 \circ \Phi_0) \Gamma_{in} z_{i, t-1}$$

where $\Gamma_{in}$ is the linear embedding layer, and $\Gamma_{out}$ is the output linear layer that maps vectors from embedding space to returns space.

Following the methodology of Gu et al., we use a standard autoencoder for the factor specification. The mathematical formulation of the factor returns is specified as the following:

$$f_t = W^0 r_t$$

which is essentially a one layer neural network without bias term and activation function. The choice of this architecture is based on the economic interpretation of the factor returns: they are themselves the return of a certain characteristic portfolio (and therefore their returns are linear combinations of the returns of the underlying assets)

In practice, the number of assets (N) could be astronomical, which would significantly increase the size and computational complexity of the factor network,

therefore we employ a dimension reduction before feeding it to the neural network:

$$x_t = (Z_{t-1}^T Z_{t-1})^{-1} Z_{t-1}^T r_t$$

the $x_t$ is the solution to the linear regression equation: $r_t = Z_{t-1} x_t + \epsilon_t$ . It can be interpreted as the observable characteristic factors return, where the factor exposures are exactly $Z_{t-1}$. This architectural choice greatly reduced the size of the factor network, and connected the factor returns to characteristics based portfolios.

It is quite common that we would end up with a characteristic matrix that is rank deficient, especially if the characteristics include "country" or "industry" loadings. To sidestep this issue, we used a linear regression with L2 regularization (Ridge regression):

$$x_t = (Z_{t-1}^T Z_{t-1} + \lambda I)^{-1} Z_{t-1}^T r_t$$

where λ is the regularization strength. The particular choice of λ is decided using the validation period. In principle the observable factor estimates obtained this way are no longer unbiased. However, we believe that this issue is trivial because the flexibility of the neural network can easily accommodate this.

Combining the two networks, the end product of our model is of the form:

$$\hat{r}_t = KAN(Z_{t-1}) W^0 x_t$$

The mean squared loss is computed using the predicted and actual returns, and then is used to jointly train our 2 networks.

## 4 Experiment

### 4.1 Experimental Setup

We have the first 30 years of data as training period (1957 - 1987), next 12 years as validation period (1987 - 1999), and all the remaining years as test period (2000 - 2016). The neural network is trained with the first 12 years of training data, and then recursively refitted with the rest of the training data, each time we extend the training sample by one year. In each refitting, the hyperparameters are tuned using the validation period, therefore it indirectly serves as input to our model. We roll it forward by one year in each refitting to maintain the same size (12 years) for the validation period.

### 4.2 Performance Metrics

The evaluation framework hinges on a robust selection of performance metrics tailored to assess the model's effectiveness in financial forecasting and risk management.

Following the benchmark set by KPS, we use total R2 and predictive R2 to evaluate model performance. In addition, we evaluate its performance in economic terms by calculating the sharpe ratio of long-short portfolios formed using the predictions made by the models.

Total R2: This metric serves as a benchmark for evaluating the goodness-of-fit of the model to the observed data. Total R2 quantifies the explanatory power of current factor realization, and therefore can be used to assess how accurate the model is to assess an asset's aggregate riskiness.

Predictive R2: This metric serves as a complement to the total R2, in that it assesses the accuracy of the prediction of the cross-sectional asset excess returns made by the model. It quantifies the model's ability to explain variation in risk compensation.

Sharpe Ratio: As a measure of risk-adjusted return, the Sharpe ratio evaluates the excess return generated per unit of risk taken by the investment strategy. The shape ratio of the long-short portfolio formed using the model's predictions quantifies the economic utility that the model has in actual applications.

## 5 Results and Discussion

The empirical findings demonstrate the KAN-based autoencoder's superior performance in financial forecasting tasks compared to traditional and previous autoencoder models. As shown in Table 1, the KAN-CA (KAN based Conditional Autoencoder) model achieves improved R² scores of 11.02%, 11.26%, and 11.32% for 1-, 3-, and 6-factor models respectively, indicating a more precise fit to the underlying data compared to Fama-French (FF) and conditional autoencoder (CA) models. Notably, the KAN-CA model's performance remains consistent across different numbers of factors, suggesting its robustness in handling varying levels of complexity in financial data. The FF model's negative R² scores across all factor models highlight the significant improvement offered by both CA and KAN-CA approaches in capturing the underlying patterns in financial markets. The model's enhanced predictive power is further evident in Table 2, where the KAN-CA model exhibits superior predictive R² scores of 0.203%, 0.203%, and 0.214% for 1-, 3-, and 6-factor models, outperforming both FF and CA models across all factor scenarios. This consistent outperformance in out-of-sample predictions underscores the KAN-CA model's ability to generalize well to unseen data, a crucial attribute for practical applications in financial forecasting. Interestingly, while the CA model shows a slight decline in performance as the number of factors increases (from 0.202% to 0.168% to 0.188%), the KAN-CA model demonstrates improved performance, particularly in the 6-factor model, suggesting its capability to effectively leverage additional factors for enhanced predictive accuracy.

Furthermore, Table 3 illustrates the model's improved risk-adjusted returns, with the KAN-CA model consistently delivering higher Sharpe ratios (0.86, 0.86, and 0.96 for 1-, 3-, and 6-factor models) compared to the CA model (0.84, 0.87, and 0.91 respectively). This improvement in Sharpe ratios indicates that the KAN-CA model not only enhances returns but does so with a more favorable risk-return tradeoff, a key consideration for practical portfolio management. The notable increase in the Sharpe ratio for the 6-factor KAN-CA model (0.96) compared to its CA counterpart (0.91) further emphasizes the model's ability to effectively utilize multiple factors for improved risk-adjusted performance. The training performance of the KAN model is visualized in Figures 2, 3, and 4, which depict the training versus validation loss for the 1-factor, 3-factor, and 6-factor models respectively. These graphs demonstrate the model's learning progression and its ability to generalize to unseen data. In all three figures, we observe a consistent decrease in both training and validation loss over the epochs, indicating effective learning without overfitting. The close alignment between training and validation curves suggests good generalization capabilities, which is crucial for reliable financial forecasting.

When compared to the MLP model results shown in Figures 1, 2, and 3, the KAN-based model exhibits improved learning stability and generalization capabilities. The MLP models, while showing decreasing loss trends, display more volatility in their validation curves, particularly in the 3-factor and 6-factor models (Figures 2 and 3). This contrast highlights the KAN model's superior ability to maintain stable performance across increasing model complexity. The smoother convergence of the KAN model, especially in the later epochs, suggests its potential for more reliable and consistent predictions in real-world financial applications. Overall, the KAN-based model's ability to capture complex, non-linear relationships within financial data is evidenced by its significant improvements in prediction accuracy across different factor models. This capability is crucial for navigating dynamic market conditions and identifying nuanced patterns that drive investment performance. The model's consistent outperformance in R² scores, predictive accuracy, and risk-adjusted returns, coupled with its stable learning behavior across various factor models, positions it as a promising advancement in financial modeling. These results suggest that the KAN-based approach not only enhances the accuracy of financial forecasts but also provides a more robust and reliable framework for decision-making in complex financial markets.

Table 1: Comparison of % R2 Scores

| Model | 1 factor | 3 factors | 6 factors |
|---|---|---|---|
| FF | <0 | <0 | <0 |
| CA | 11.06 | 11.39 | 11.29 |
| KAN CA | 11.02 | 11.26 | 11.32 |

Table 2: Comparison of % Predictive R2 Scores

| Model | 1 factor | 3 factors | 6 factors |
|---|---|---|---|
| FF | <0 | <0 | <0 |
| CA | 0.202 | 0.168 | 0.188 |
| KAN CA | 0.203 | 0.203 | 0.214 |

Table 3: Comparison of Sharpe Ratio of models

| Model | 1 factor | 3 factors | 6 factors |
|---|---|---|---|
| CA | 0.84 | 0.87 | 0.91 |
| KAN CA | 0.86 | 0.86 | 0.96 |

**6 Conclusion**

This research introduces a KAN-based autoencoder latent factor model that improves upon traditional asset pricing models by capturing non-linear relationships. The empirical analysis demonstrates the model's superior predictive power and economic interpretability.

The findings have significant implications for financial economics, providing a more accurate and interpretable model for asset pricing. This approach can be extended to other financial applications, such as risk management and portfolio optimization.

The study is limited by the advancement of KAN model architecture. Unlike MLP, whose training behavior, regularization technique, choice of layers / activation functions are thoroughly studied over the past decades, KAN is still a relatively novel invention whose capacity remains largely unexplored. There may exist better architecture for KAN networks to approximate the beta function. We leave this engineering problem for future research.

Also, one of the key advantages of KAN over MLP is its interpretability. Through the exploration of spline function in each KAN layer, we can visualize how each covariates of assets interact with their betas to each latent factor. This visualization may provide more information of how the covariates affect the risk aggregation of assets, and may further simplify the model.

Future research could explore extending the model to other asset classes and incorporating additional covariates. Additionally, further advancements in KANs and other deep learning techniques could provide even greater improvements in model performance.

**REFERENCES**

[1] Eugene F. Fama and Kenneth R. French. 1993. Common risk factors in the returns on stocks and bonds. Journal of Financial Economics 33, 1 (1993), 3-56.

[2] Robert Novy-Marx. 2013. The Other Side of Value: The Gross Profitability Premium. Journal of Financial Economics 108, 1 (2013), 1-28.

[3] Clifford S. Asness, Andrea Frazzini, and Lasse H. Pedersen. 2019. Quality Minus Junk. Review of Accounting Studies 24, 1 (2019), 34-112.

[4] William F. Sharpe. 1964. Capital Asset Prices: A Theory Of Market Equilibrium Under Conditions Of Risk. The Journal of Finance 19, 3 (1964), 425-442.

[5] Gary Chamberlain. 1983. Funds, Factors, and Diversification in Arbitrage Pricing Models. Econometrica 51, 5 (1983), 1305–1323.

[6] Shihao Gu, Bryan Kelly, and Dacheng Xiu. 2021. Autoencoder asset pricing models. Journal of Econometrics 222, 1 (2021), 429-450.

[7] Ziming Liu, Yuanqi Wang, Shuqi Vaidya, Frederik Ruehle, James Halverson, Marin Soljačić, and Max Tegmark. 2024. KAN: Kolmogorov-Arnold Networks. arXiv preprint arXiv:2404.19756 (2024).

[8] Ziming Liu, Yuanqi Wang, Shuqi Vaidya, Frederik Ruehle, James Halverson, Marin Soljačić, and Max Tegmark. 2024. KAN: Kolmogorov-Arnold Networks. arXiv preprint arXiv:2404.19756 (2024).

[9] Andrey N. Kolmogorov. 1957. On the representation of continuous functions of many variables by superposition of continuous functions of one variable and addition. Doklady Akademii Nauk SSSR 114, 5 (1957), 953-956.

[10] Ming-Jun Lai and Zuowei Shen. 2021. The Kolmogorov superposition theorem can break the curse of dimensionality when approximating high dimensional functions. arXiv preprint arXiv:2112.09963 (2021).

[11] Yifan Duan, Lantian Wang, Qin Zhang, and Jing Li. 2022. FactorVAE: A Probabilistic Dynamic Factor Model Based on Variational Autoencoder for Predicting Cross-Sectional Stock Returns. In Proceedings of the AAAI Conference on Artificial Intelligence, Vol. 36. 4468-4476.

[12] Eunbi Kim, Taehyun Cho, Baeho Koo, and Hoon Gyu Kang. 2023. Conditional autoencoder asset pricing models for the Korean stock market. PLOS ONE 18, 7 (2023), e0281783.

[13] Kevin Huynh and Guillaume Lenhard. 2022. Asymmetric Autoencoders for Factor-Based Covariance Matrix Estimation. In Proceedings of the Third ACM International Conference on AI in Finance (ICAIF '22). Association for Computing Machinery, New York, NY, USA, 403–410.

[14] Shihao Gu, Bryan Kelly, and Dacheng Xiu. 2021. Autoencoder asset pricing models. Journal of Econometrics 222, 1 (2021), 429-450.

# APPENDIX

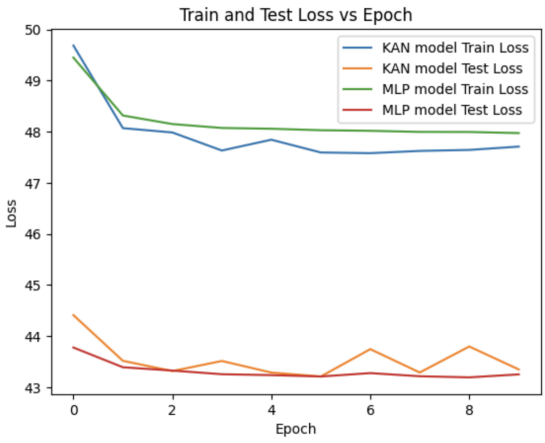

Figure 2: 1 factor model Training vs Validation Loss

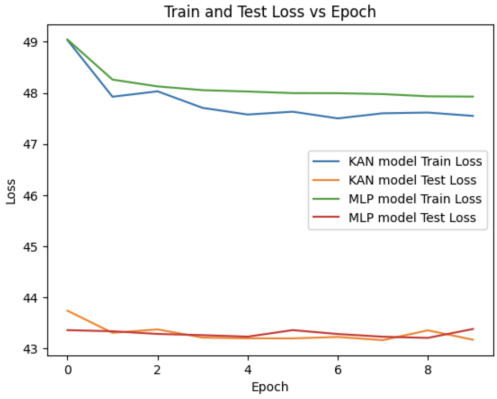

Figure 3: 3 factors KAN Training vs Validation Loss

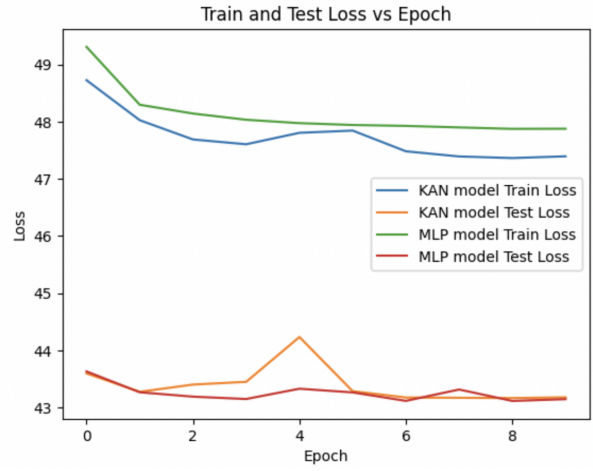

Figure 4: 6 factors KAN Training vs Validation Loss